# Aligning Robot's Behaviours and Users' Perceptions Through Participatory Prototyping

Pamela Carreno-Medrano[1], Leimin Tian[1], Aimee Allen[1], Shanti Sumartojo[2], Michael Mintrom[3], Enrique Coronado[4], Gentiane Venture[4], Elizabeth Croft[1] and Dana Kulić[1]

*Abstract*— Robots are increasingly being deployed in public spaces. However, the general population rarely has the opportunity to nominate what they would prefer or expect a robot to do in these contexts. Since most people have little or no experience interacting with a robot, it is not surprising that robots deployed in the real world may fail to gain acceptance or engage their intended users. To address this issue, we examine users' understanding of robots in public spaces and their expectations of appropriate uses of robots in these spaces. Furthermore, we investigate how these perceptions and expectations change as users engage and interact with a robot. To support this goal, we conducted a participatory design workshop in which participants were actively involved in the prototyping and testing of a robot's behaviours in simulation and on the physical robot. Our work highlights how social and interaction contexts influence users' perception of robots in public spaces and how users' design and understanding of what are appropriate robot behaviors shifts as they observe the enactment of their designs.

## I. INTRODUCTION

As the number of robots deployed in public spaces such as airports, hospitals, hotels, or cafes rises, the likelihood of every-day encounters and potential interactions with a robot continues to increase. Up until now, since relatively few people have interacted with a robot at all [1], it is not surprising that robots deployed in the real world may fail to gain acceptance or engage their intended users. This failure typically occurs when the robot behaviours do not match people's understanding and expectations about what a robot should be capable of and what is appropriate for a robot to do in such contexts [2], [3].

Previous Human-Robot Interaction (HRI) studies found consistent evidence that discrepancies between a robot's actual capabilities and a user's expectation of capabilities can negatively impact the user's acceptance, use of, and willingness to interact with a robot [4], [5]. Thus, researchers have investigated people's expectation about robots [6], how such expectation was formed [1] and how factors such as a robot's physical appearance [7], interaction skills and expected future role can affect such expectation [3]. However, little is known about how expectations of robot capabilities might change as people engage and interact with real robots.

Robot designers and HRI researchers often employ participatory design (PD) methods to better understand individual contextualized opinions about what behaviours and actions people think are appropriate for a robot to do. This methodology allows intended end-users of a robotic system to be actively involved in the design of the system, thus supporting the development of robot functionalities that the end-users consider to be more appropriate [8]. Recent HRI studies have demonstrated the effectiveness of PD in applications such as rehabilitation [9], teen or elderly mental health [10], [11], and support of aging users [12]. However, in most of these studies, participants are rarely shown a realisation of the robot functions they designed or interact directly with a robot. Thus, these studies fail to account for any potential change in end-users' expectations after they engage or interact with the new technology.

Motivated by these limitations, and extending our previous work [13], here we investigate how users of public spaces understand and perceive robots, what roles and uses they think are appropriate for robots in public spaces, and how these understandings and expected uses change as they engage with a real robot. To support this goal, we propose a participatory design workshop methodology in which participants are actively involved in the design and refinement of a robot's behaviour. During the workshop, participants can nominate what they personally would prefer or expect a robot to do in a given setting as well as refine their expectations as they observe tangible realisations of these behaviours. As a proof of concept, an exploratory study was conducted in which participants conceptualised and programmed the behaviours of a social robot situated in a public space of their choice. In [13], we report on the first part of the study, where robot behaviour prototyping was carried out in simulation. Here we describe the workshop with physical prototyping and investigate the causes for participants' shift in perception regarding the robot behaviours and role. We also observed how these changes in perception were reflected in the redesign of the robot behaviors.

*This project is supported by the Monash University Interdisciplinary Research Seed Grant. This research is supported by an Australian Government Research Training Program (RTP) Scholarship.

[1]Pamela Carreno-Medrano, Leimin Tian, Aimee Allen, Elizabeth Croft and Dana Kulić are with the Faculty of Engineering, Monash University, Melbourne, Australia {pamela.carreno, leimin.tian, aimee.allen, elizabeth.croft, dana.kulic}@monash.edu

[2] Shanti Sumartojo is with the Emerging Technologies Research Lab, Monash University, Melbourne, Australia shanti.sumartojo@monash.edu

[3] Michael Mintrom is with the School of Social Sciences, Monash University, Melbourne, Australia michael.mintrom@monash.edu

[4] Enrique Coronado and Gentiane Venture are with the Department of Mechanical Systems Engineering, Tokyo University of Agriculture and Technology Japan enriquecoronadodozu@gmail.com, venture@cc.tuat.ac.jp

## II. RELATED WORK

To develop robotic systems that can seamlessly interact with and work alongside humans, HRI researchers have focused on examining what factors are key in people's perception and evaluation of such systems. Multiple studies [14], [15], [5], [16] in which participants are presented with an a priori defined robot functionality, point towards the importance of matching the users' expectations regarding the functions and capabilities of a robot and the functions that the robot is actually capable of executing. Any potential gap or discrepancy between a user's expectations and the reality of a robot's capability can result in frustration, reduced trust [5] or a change in the intended use of the robotic system [3]. Thus, it is critical to understand what these expectations are and how they change as people interact with robots.

Participatory design is an approach that includes end-users in design and development process of a system [17]. PD methods facilitate the development of more nuanced and contextualized applications as they leverage the end-users' expertise and know-how [10], and also allow the designers of the system to better understand what the end-users themselves would prefer or expect from the outcome of the design in a given setting [5]. In the context of HRI, an increasing number of studies employ PD methods to design new robots and/or robotic applications. For instance, Lee et al. [12] used different collaborative methods to understand how older adults frame aging and how existing or new robots fit into the participants' view of the aging process. Deuff et al. [18] employ user-centered and participatory methods to study how young retirees perceive and envision robotics systems for the home. Winkle et al. [9] worked with therapists to design strategies and behaviours that existing social robots could use to increase patients' engagement in rehabilitation. PD has also been used to design HRI applications for mental well-being interventions [11], [10] as well as the development of educational or assistive robot tools for children [19] and adults [20] with visual impairments.

These HRI studies indicate that participatory design methodologies can be used for examining and identifying end-users' expectations regarding the uses, functions and capabilities of a robot. However, since most studies are centered on one-on-one interaction scenarios within specific populations, they offer little or no insight on the expectations the general population have about robots that are deployed in public spaces, have varied interaction goals, and will interact with a diverse user population. Moreover, although the majority of participatory design studies in HRI focus on prototyping new robot functions, end-users rarely get the opportunity to interact and test out their designed prototypes with an actual robot in context. Thus, if the users' expectations fail to be met or change as a consequence of the interaction with the real robot, this can result in lower user acceptance of the designed robot functions. Our participatory design workshop methodology allows participants to design behaviors and then test these behaviors first in a simulated environment and then on a physical robot. After observing the execution, participants can make iterative changes to improve the robot behavior design to address their observations. This allows us to investigate users' perceptions and expectations of robots and their use in public spaces as they design new behaviors and observe realisations of those behaviors both in simulation and on the physical robot.

## III. METHODOLOGY

To gain further insight as to how people understand robots in public spaces and what uses they envision for them, we proposed and implemented a participatory design workshop methodology in which small focus groups of participants actively engage in the prototyping, programming and testing of behaviours. The proposed methodology consists of two workshops, the first one in simulation and the second one with the physical robot.

Each workshop is composed of three phases. First, participants are introduced to the objectives of the workshop as well as the different tools they will use for prototyping a robot's behaviour. Next, participants are divided into smaller focus groups. Each group works on the iterative design and testing of multiple robot behaviours of their choice. Finally, the focus groups rejoin to demonstrate and explain their designs of robot behaviors. Due to COVID-19 physical distancing restrictions and to ensure the safety of our participants and researchers, an online format was adopted for both workshops. In the following sections, we detail each workshop and elaborate further on the rationale for the choices made in the design and planning of these workshops.

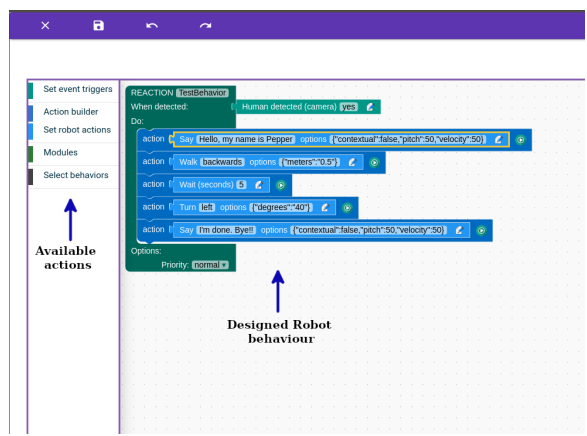

Fig. 1: The RIZE end-user programming interface.

### A. Tools

In contrast to other participatory design methodologies in which paper sketching and storyboarding are the main tools employed by participants during the design process [11], we provided participants with an end-user friendly robotic programming interface instead. This interface allows participants to immediately observe what a specific action or combination of actions would look like when executed by the robot, facilitating a better sense of the robot's functionalities and capabilities.

We used Robot Interface from Zero Expertise (RIZE) [21] as our end-user programming interface. RIZE (see Fig. 1) is a modular and distributed end-user development framework that enables non technically skilled people to create robotic applications. RIZE follows a component-based methodology, in which all robot action and perception functions are encapsulated as visual building blocks that a user can combine through simple logic structures. The same programming interface is used during both workshops.

To support the programming interface and help participants navigate the various robot functions offered for the workshop, two additional "programmming" resources are provided to the participants. First, during each workshop, participants are accompanied by a technical expert, a member of the research team, whom we refer to as the *translator*. The translator is there to assist with the implementation of the behaviours proposed by the participants and answer participants' questions about the robot and the programming interface. Second, each participant is provided with a *cheat-sheet* list with all available robot actions.

As the robotic platform for our workshops, we chose Pepper. Pepper is one of the most popular existing social robotic platforms and is frequently deployed in public spaces [22]. Additionally, its accompanying simulator Choregraphe [23] offers an easy and intuitive way for participants to test their designed behaviours and monitor the physical robot's status.

### B. Workshop One: Behaviour Design in Simulation

The first workshop in the proposed participatory design methodology is run using the simulated environment Choregraphe [23]. The objective of this workshop is to allow participants to familiarize themselves with the different actions Pepper can execute and the logic and programming structure that are used to design behaviours in which multiple actions are combined. It also provides us with a general view of participants' initial expectations and understanding of robots.

This workshop consists of three main phases. First, participants join an open discussion on what they think is a robot and their prior experiences with robots in public spaces. Second, participants are randomly assigned to smaller focus groups for the participatory prototyping session. Each group is tasked with designing a set of robot behaviors for a public space of their choosing with the support of RIZE, the translator, and the cheat-sheet. Finally, all groups rejoin to demonstrate their designs of robot behaviors and explain the rationale behind their designs. Detailed description and analysis of this first workshop can be found in [13].

### C. Workshop Two: Behaviour Refinement with a Physical Robot

The second workshop is conducted with the physical robot and has two main objectives: *1.)* allow participants to observe Pepper's behaviours in a physical setting, and *2.)* support the design of interactive robot behaviours that cannot be achieved in a simulated environment. This workshop is intended as a continuation of the first simulation-only workshop and thus requires the same participants to be recruited. Prior to the workshop, participants are asked to answer a set of validated questionnaires to assess their initial expectations and perceptions of the robot. The questionnaires are: *i.)* the GodSpeed questionnaire [24] for likeability and perceived intelligence, the Trust questionnaire [25], and the Reliability questionnaire [26]. An adhoc questionnaire that measures how successful participants expect the robot to be at executing a designed behavior and achieving its intended outcomes is also included.

After being introduced to the physical robot, participants engage anew in a one-hour participatory prototyping session. Participants are randomly separated into small focus groups as done in the first workshop. During this second programming session, each group has remote access to a Pepper robot and performs iterative changes to the behaviours designed during the first workshop. After each iteration, participants can watch the execution of the edited behavior on the physical remote robot through a live-feed of the robot and through robot status monitoring in Choregraphe (see Fig. 2). When the designed behaviours include an interaction with a user, one of the members of the research team plays the role of the interactee.

Finally, all groups rejoin for demonstration and discussion. At the end of this second workshop, participants are asked to answer the same set of pre-workshop questionnaires. Additionally, participants are also asked to elaborate on their perception of the robot's execution of the designed behavior and the improvements needed for the robot to be more useful and acceptable in public spaces.

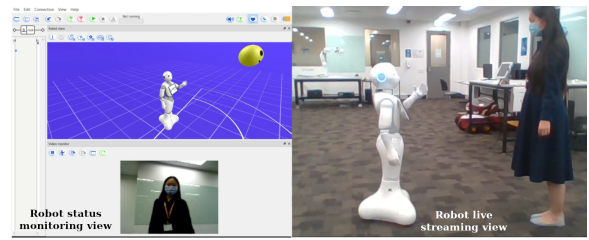

Fig. 2: Robot monitoring interface and live-streaming of physical robot.

### D. Participants and Workshops Implementation

A total of 12 participants (5 males, 7 females) recruited from Monash University staff members took part in both workshops. Participants came from several faculties including architecture, arts, criminology, among others, and most of them had no prior experience with robots or programming. They attended both workshops remotely via a video conferencing platform (Zoom) and provided written consent for their participation and for recording of the sessions. One participant was unable to attend the second workshop due to technical difficulties, leaving 11 participants for the second workshop. The study protocol was reviewed and approved by the Human Research Ethics Committee of Monash University.

Each workshop took approximately two hours, with the second workshop taking place one month after the first one. In both workshops, participants were divided into 2 focus groups, hereinafter referred to as *group one* and *group two*. During the first workshop, each group agreed on a public space setting for Pepper and designed multiple behaviours associated to this setting. *Group one* envisioned Pepper at a shopping mall and designed behaviours such as greeting people, a social distancing reminder, and an idle behaviour that randomly displays entertaining gestures such as dancing. *Group two* placed Pepper at a tourist attraction and designed behaviours such as cleaning the floor, providing directions and taking pictures for tourists.

Coincidentally, both groups designed a behaviour in which Pepper attends to a lost child. This common behaviour was used as the starting point of the second workshop. During this workshop, when required during robot behaviour testing, a member of the research team took the role of the lost child. After observing the robot's performance, participants made adjustments that they deemed necessary to their initial designed behaviour. Each group was assigned their own *translator* for the duration of both workshops.

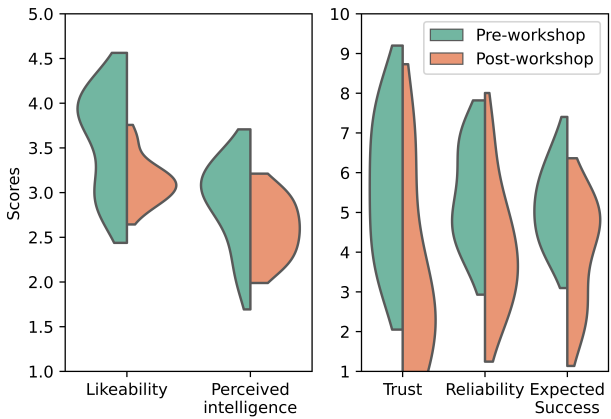

Fig. 3: Distribution of scores for likeability, perceived intelligence, trust, reliability and expected success before (green) and after (orange) workshop two.

## IV. ANALYSIS AND RESULTS

We use a mixed qualitative and quantitative approach to analyze: *1.)* how the proposed participatory prototyping methodology changes participants' understanding and expectations about Pepper, *2.)* how this change in perspective is reflected in the robot behaviors designed by the participants, and *3.)* what uses participants think are appropriate behaviours for Pepper in public spaces. This analysis is centered on the recordings and questionnaires obtained during the second workshop.

### A. Questionnaires

To determine if there has been a change in participants' perception and expectations of Pepper, we compare the ratings obtained from the questionnaires participants answered before and after the second workshop. Based on a participant's responses, a composite score for each element; likeability, perceived intelligence, trust, reliability and expected success was calculated by averaging the scores the participant assigned to all items for that category. Pre- and post-workshop ratings for all categories are shown in Fig. 3.

Since some of our questionnaire data violated the normality assumption required for a parametric $t$-test, non-parametric procedures were used instead. Similarly, after validating that our data is non-symmetric, we decided to use a two-sided paired sign-test. This test determines whether the median difference between the participants ratings before and after the second workshop is not equal to zero (*alternative hypothesis*). Although Fig. 3 visually indicates an overall decrease in all ratings for most of our participants, the paired sign-test found no statistically significant rating decrease ($p-$value $> 0.05$) for likeability nor perceived intelligence. However, the perceived trust, reliability, and expected success ratings obtained after the workshop showed a statistically significant median decrease ($p-$value $< 0.05$ for all categories) compared to the ratings obtained prior to the second workshop.

### B. Inductive Coding of Recordings

The pre- and post-workshop questionnaires indicate that a significant change in participants' perceptions and expectations of Pepper happened during the participatory prototyping session. To understand what circumstances encountered during the workshop lead to this change in perception, we use inductive coding on the workshop recordings.

After the workshop recordings were transcribed, the participants' comments regarding their perceptions and opinions of the observed robot behaviour as well as their reasoning for proposing a change in their behaviour prototypes were grouped into categories using inductive coding. The coding was done by the members of the research team that played the role of *translators* during the workshop. We used approximately $60\%$ of the data to compute inter-coder reliability using the "Kappa" coefficient. The results indicate a substantial agreement ($\kappa = 0.64$, $95\%$ confidence interval $[0.59, 0.70]$) according to the interpretation given in [27].

Based on this inductive coding scheme, we identified five types of circumstances that participants encountered during the workshop that can potentially explain the changes in perception observed in the questionnaires. A description of each type of situation is given in TABLE I. We note that these categories are not exclusive, and that a similar situation could fall into more than one category.

During the participatory prototyping session in the second workshop, when asked to elaborate on their reasoning and decisions regarding the changes in designed robot behaviour, participants referred to the above-mentioned categories with the frequency shown in Fig. 4. These frequency percentages were calculated by averaging the annotators' label counts.

From Fig. 4 we observe that although failures and limitations in the tools offered to participants during the work-shop (e.g., the programming interface and the functionalities implemented in Pepper) account for approximately 11% of participants' perceptions regarding Pepper's (in)capability to fulfill the role and behaviour participants assigned to it (i.e., attend to a lost child), most of the participants' answers and comments alluded to the complexity of the application context chosen for this workshop (i.e., public spaces) ($\sim 50\%$), the gap between the designed behaviour and the expected outcome ($\sim 24\%$), and the risk Pepper can pose to potential interactees ($\sim 15\%$).

TABLE I: Description of the five types of circumstances encountered by participants during the 2nd. workshop and used to explain the changes in perception of the robot seen in participants.

| Trigger | Description | Associated Codes (Most Frequent to Least Frequent) |
|---|---|---|
| Context complexity | All circumstances in which a behaviour initially intended for the robot was later found to be unsuitable for any robot at all (e.g., robots lack the nuanced judgement needed to handle a delicate situation such as a lost child and this should be up to a person to decide instead) or the robot's actions seemed to be overlooking the social and physical context in which the interaction with a human could take place (e.g., an initially thought benign robot action can be seen as disconcerting to a child). | 1) User context<br>2) Social context<br>3) Unsuitable goal<br>4) Spatial context<br>5) Physical context |
| Behaviour limitations | All circumstances in which although the robot's actions were executed without fault, the outcome was seen as unfitting, unexpected, inconsistent, lacking flexibility or following an inappropriate pace (e.g., the robot was executing a single script and seemed incapable of handling answers or behaviours other than what was expected). In this category we also include the situations in which participants questioned whether a given robot behaviour would generalize to a different context or group of users | 1) Inappropriate behaviour<br>2) Generalize<br>3) Unexpected behaviour<br>4) Nonadaptive behaviour<br>5) Inconsistent behaviour |
| Risk concerns | All circumstances in which the robot's actions were perceived as problematic due to potential safety, ethical or legal issues (e.g., robot makes decisions based on potentially faulty emotion recognition software or robot moves in an abrupt way and people could get hurt) | 1) Ethical concern<br>2) Safety concern<br>3) Liability concern |
| Performance and interaction failures | All circumstances in which there was a technical failure in the robot's function (e.g., the speech recognition functionality fails to recognize a key term) or the designed behaviour was executed without fault but from the participants' perspective fails to engage the intended interactee or to address the interaction flow properly (e.g., the robot was perceived as running through the motions, without pauses in between actions and no real interaction) | 1) Performance failure<br>2) Interaction failure<br>3) Engagement failure |
| Tools limitations | All circumstances in which both the robot was perceived as limiting for its intended functions (e.g., the robot's physical appearance and aesthetics are not designed for the task at hand) or the programming interface was missing a functionality that will make the robot's behaviour more flexible or less inconsistent (e.g. a more complex dialogue system) | 1) Interface limits<br>2) Robot design limitation |

Regarding the context complexity category, after seeing Pepper execute the lost child behaviour, participants started questioning whether Pepper or any other robot has the capacity to navigate through the intricate sequence of decisions necessary to handle this situation in a sensible and reliable way. Similarly, they reflected on the unpredictability of the intended interactee as well as the robot's capacity to determine whether the situation was handled correctly or a special set of rules should be applied (e.g., a participant questioned whether some robot actions might be disconcerting to a child with autism). Participants also commented on the likelihood of the sensing capabilities of the robot failing once it is deployed in uncertain and dynamic environments such as public spaces and how these errors should be handled when the robot's decisions depend on the accuracy of its sensing.

As for behaviour limitations, most participants' observations were around how Pepper's behaviour lacked the flexi-

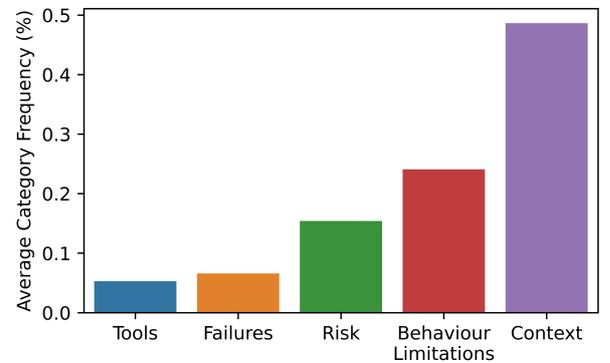

fig. 4: Categorization of participants' responses regarding their perceptions of the robot and their decisions to make adjustments to robot behaviours. Bars indicate the frequency of each category as averaged between the two annotators.

bility and adaptability required for a fluid, seamless interaction. Participants commented on how it seemed that Pepper was running through the different actions programmed by them without taking into consideration the actions of the interactee or the interactee's responses to the verbal prompts the robot was executing. Participants also highlighted that the robot's movements were abrupt, aggressive and intrusive and thus made them doubt the safety of the people around the robot could be ensured. This observation might explain the decrease in the trust ratings observed at the end of the workshop. Some participants questioned if it was fitting for Pepper to have the role of initiator or whether it should be assigned a more passive role and wait for a person to express their wish to interact with the robot

*C. Adjustments to Robot Behaviour*

Three types of risks were often mentioned by the participants when explaining the adjustments they made to the robot's behaviour. First, participants were concerned that in the role of attending to a lost child, Pepper lacked the physical, observable cues that allowed it to present itself as a safe person (e.g., wearing a uniform), i.e, someone that is safe for children to have a conversation with or be approached by. Second, since some of Pepper's perception capabilities rely on functionalities such as face detection and facial expression recognition, some participants voiced their concerns about the biases of such systems [28] and whether the non-threatening appearance of Pepper might mislead the users of public spaces about the potential harm of such systems. Finally, participants also reflected on how much responsibility should be attributed to the robot and to the interactee in the case of an issue.

Besides the comments and opinions participants shared during the workshop, the participants made changes to Pepper's designed behaviours as they gained a better understanding of Pepper and its capabilities. On the one hand, after quickly deciding that it was not appropriate for Pepper or any other robot to attend to a delicate and complex situation such as lost child, *group one* chose to redefine Pepper's uses to less delicate situations (e.g., provide directions to an adult lost in the shopping mall). Similarly, they agreed on limiting the agency of the robot and make it a passive helper that waits to be approached by any potential user. On the other hand, after noticing the lack of responsiveness and flexibility, and the unpredictability of Pepper's actions in their initial implementation of the lost child behaviour, *group two* tried to incorporate more pauses between the different actions Pepper would execute as well as sentences that indicated to the interactee what action Pepper would execute next. Similarly, participants in this group also aimed at improving Pepper's decision process so as to better reflect the complexity of a lost child approaching or being detected by the robot.

This iterative behaviour refinement process also allowed participants to increase their awareness of the intricacy and difficulty involved in the design of appropriate interactive behaviours as well as the fact that despite their best efforts, robots are still fallible and have limitations.

## V. CONCLUSION AND DISCUSSION

In this paper we have proposed a novel methodology for understanding user's expectations of robots situated in complex interaction contexts. In comparison to existing participatory design methodologies, our novel online participatory prototyping workshops enabled the participants to actively test and refine their design of a robot's role and behaviors with an end-user-friendly robotic programming interface. With this proposed methodology, we studied how participants' perceptions of robots' capabilities and their appropriate uses in public spaces change as they engage and interact with a robot. This change in perception was reflected in the decline of perceived likeability and intelligence ratings and significantly, trust, reliability and expected success scores. Our observations are supported by previous studies where users' evaluations decreased after interacting with a robot for several minutes [29], witnessing how a robot's interactive capabilities fail to conform to users' expectations [16], or observing a robot act in a faulty or unexpected manner [30]. We note that the proposed hands-on, behaviour prototyping methodology allowed participants to encounter all these situations in the span of one hour.

In contrast to existing robot-centered approaches focused on improving a robot's functionalities to fulfill users' expectations [15], our user-centered study confirms that contextual factors have far more significant impact on a user's perceptions and expectations of a robot. We found that the robot's (in)ability to respond to and account for the social and interactive context and complexity inherent in interaction settings was the biggest limitation identified by the participants and the major rationale behind their design decisions. This observed relation between the level of interaction skills and people's evaluation and perception of a robot's competence and suitability for a role such as the one participants initially envisioned for Pepper (i.e, attending to a lost child) is also supported by previous studies [2], [3].

The proposed methodology also provides further insight on how the changes in participants' perceptions and expectations of a robot are reflected in the modifications and refinements they introduced to robot behaviours during the workshop. While one group decided to keep the attending to a lost child role and focused on improving Pepper's decision process, awareness and responsiveness to the user's potential actions, the other group decided on redefining Pepper's uses to less delicate situations (e.g., provide directions to an adult lost in the shopping mall) and limiting the agency of the robot by making it a passive helper.

The observations made during our exploratory study provide us with a first indication of the efficacy of such workshops in analyzing users' perception and expectation of robots in complex contexts such as public spaces. More importantly, we demonstrate how by allowing participants to quickly nominate, program, and test the actions and behaviours they think are appropriate for a robot we can elicit and capture changes in perception due to factors such as technical failures, limitations in robot's capabilities, and more abstract constructs such as moral and ethical concerns. As directions for future work, we plan to investigate a compressed format of both workshops so as to relax the requirement on retention of the same participant group. Similarly, we plan to extend our observations to different classes of robots and, whenever possible, to conduct in-person workshops in which participants share the same space with the robot and have direct interactions with it.